# Augmented Q Imitation Learning (AQIL)

Xiao Lei Zhang
York University
*Toronto*, Canada
zhang205@cse.yorku.ca

Anish Agarwal
University of Waterloo
*Waterloo*, Canada
a22agarw@outlook.com

The study of unsupervised learning can be generally divided into two categories: imitation learning and reinforcement learning. In imitation learning the machine learns by mimicking the behavior of an expert system whereas in reinforcement learning the machine learns via direct environment feedback. Traditional deep reinforcement learning takes a significant time before the machine starts to converge to an optimal policy. This paper proposes Augmented Q-Imitation-Learning, a method by which deep reinforcement learning convergence can be accelerated by applying Q-imitation-learning as the initial training process in traditional Deep Q-learning.

**Reference code:** https://github.com/veda-s4dhak/AQIL

*Keywords — Q-imitation-learning, Deep Reinforcement Learning, Deep Q-learning, Behavioral Cloning, SQIL*

## I. INTRODUCTION

Imitation learning in deep learning systems is generally either focused on modeling the behavior of an expert system (behavioral cloning) or focused on modeling the reward function which best approximates the expert system's behavior (inverse reinforcement learning).

The performance of an imitation learning system alone is limited by the performance of the expert player. In the ideal sense we want systems which can increase the upper bound of performance by going beyond that of an expert system. To achieve this, imitation learning alone is not sufficient. We are bounded by the limits of supervision.

Current fully unsupervised deep learning systems which have performance beyond known expert systems have been designed using reinforcement learning, where machines are learning from direct environment interaction. Deep reinforcement learning[6] (DQN) however requires quite a bit of training period till the model reaches expert level performance. This is exacerbated in increasingly complex environments.

It seems that there is a mutual advantage to merge reinforcement learning with imitation learning. A reinforcement learning model can accelerate its initial training time by imitating an expert system. An imitation learning model can increase its upper bound and go beyond the expert system by switching to direct environment interaction.

In this paper we consider this augmentation. We use traditional imitation learning approaches as a precursor to deep reinforcement learning. A deep neural network first imitates an expert system and then is allowed to reinforce directly through the environment. We first setup the framework of the experiment, followed by the implementation details and concluding with the experimental results.

The imitation learning framework we used is a custom Q-imitation-learning protocol similar to SQIL[1]. Whereas SQIL uses a soft Bellman equation[5], we use the Bellman equation with a hard argmax. Q imitation learning is the same as traditional Q-learning in all aspects except that the state-action reward is determined by adherence to an experts state-action rather than from direct environment feedback. This paper uses a Gaussian reward function proportional to difference between expert's action and agent's action.

We follow the Q-imitation-learning with a traditional Q-learning training sequence. In the Q-learning portion we use a Gaussian reward function proportional to the difference between the optimal state and the actual state.

## II. FRAMEWORK

### A. Markov Decision Processes

We define the MDP parameters as {S, A, P, R, I}, where:

- S is the set of states
- A is the finite set of actions
- P = P(s, a, s') is the state transition probability which denotes the probability to transition to state s' given than the previous state was s and action a was taken.
- R = R(s, a) is the reward in state s given action a was taken
- I is the initial state distribution

Additional parameters are specified as follows:

- N denotes the number of episodes
- T denotes the time horizon
- $\pi$ denotes the policy that determines which action is taken at state s
- $\pi'$ denotes the experts policy
- $\pi''$ denotes the optimal policy

## B. Problem Definition

Given the above we can make the following conclusions:

- $V(\pi) = T * E\ R(s, \pi(s))$

  denotes the total rewards of all trajectories given the initial state I

- $V(\pi') - V(\pi)$ denotes the imitation regret

- $V(\pi'') - V(\pi)$ denotes the reinforcement regret

- $V(\pi'') - V(\pi')$ denotes the expert regret

The goal of augmented reinforcement learning is to accelerate reduction of reinforcement regret to the point where it is below expert regret. That is using imitation learning to reach the point where

$$V(\pi'') - V(\pi) \leq V(\pi'') - V(\pi')$$

as fast as possible.

## III. IMPLEMENTATION

### A. Agent-Environment Interaction

We implement our experiment in CartPole-v1 Gym environment from the OpenAI gym. CartPole is a conventional controls problem and is suitable for imitation learning since an expert model is readily available in the form of a PID controller.

The mechanics of CartPole consist of a pole attached by a joint to a cart, which is controlled by applying a force of +1 or -1. The pole initially starts upright and the goal is to prevent the pole from falling over. Each episode ends, when the pole is more than theta degrees from the vertical or the cart moves more than 2.4 units away from the center. For this experiment, we set theta to 50 degrees to reduce learning time of each agent.

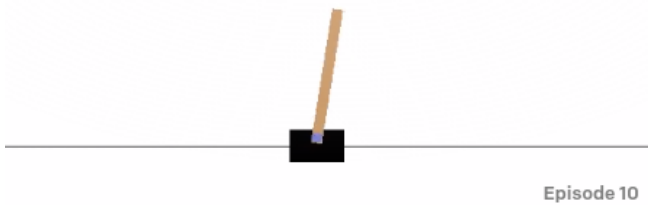

Figure 1. View of Cartpole-v1 environment.

We first train the model via Q-imitation-learning by modeling the PID. Then we train the model using deep reinforcement learning directly from the environment. We compare these results to a model trained via deep reinforcement learning alone and to a model trained via imitation learning alone.

In both Q-imitation-learning and deep reinforcement learning, we used the Q-learning methodology for training. During the imitation learning process, the Q-learning model optimizes the reward based on following the expert input.

The expert input was taken by implementing a simple PID controller to control the cart. The PID system was tuned to score much higher than an average human player. The proportional, integral and derivative parameters are as follows:

$$P = 0.6, I = 0.00625, D = 0.8$$

The reward during Q imitation learning is a Gaussian function defined below. The reward depends on the difference between the expert action and the model action as well as the difference between the optimal and actual pole angles. The reward function is highest when the pole angle is optimal and the model action matches the expert action.

$$R(\theta, a_{PID}, a_{model}) = 0.2 e^{-\frac{1}{2}\left(\frac{\theta_{optimal} - \theta}{\sigma_1}\right)^2} + 0.8 e^{-\frac{1}{2}\left(\frac{a_{PID} - a_{model}}{\sigma_2}\right)^2}$$

$\theta_{optimal} = 0 \wedge \sigma_1 = 10 \wedge \sigma_2 = 0.5$

For deep reinforcement learning the Gaussian reward function becomes

$$R(\theta) = e^{-\frac{1}{2}\left(\frac{\theta_{optimal} - \theta}{\sigma_1}\right)^2}$$

$\theta_{optimal} = 0 \wedge \sigma_1 = 10$

This reward function is based on the difference between the target and actual pole angles

We define the loss function for the $i^{th}$ training step as the Bellman error.

$$L_i(\theta_i) = (y_i - Q(s, a; \theta_i))^2$$

where

$$y_i = r + \gamma \max_{a'} Q(s', a'; \theta_{i-1})$$

### B. Model Architecture

The model architecture is shown in Figure 1. We use a fully connected neural network model for the Deep Q-Learning and Imitation Learning agents. We use ReLU activation for the inner layers and linear activation for the output layer.

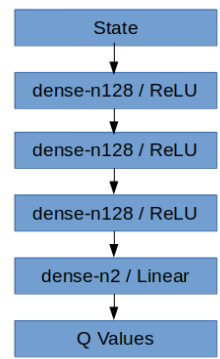

Figure 1. DQN Model Architecture. Dense layers are described by number of features (n).

## C. Imitation Training Methodology

For imitation training, we use the forward training methodology used by Ross and Bagnell (2010)[2] modified to use a stationary policy. We use a stationary policy since the T is unbounded in the training environment.

The training methodology is summarized as follows:

---
**Algorithm 1:** Q Imitation Training

1:   Initialize $\pi$
2:   **For** I = 1 to num_epochs **do**
3:       Execute x trajectories using $\pi'$
4:       Sample dataset D = {states, action} taken by expert
5:       Train $\pi$ using DQN with $Reward = R(\theta, a_{PID}, a_{model})$
6:   **End for**
7:   **Return** $\pi$
---

## D. Reinforcement Learning Methodology

The reinforcement training is similar to imitation training except that the training classifier uses a reward function directly from the environment rather than the expert player.

---
**Algorithm 2:** Reinforcement Training

1:   Initialize $\pi$
2:   **For** I = 1 to num_epochs **do**
3:       Execute x trajectories using $\pi'$
4:       Sample dataset D = {states, action} taken by expert
5:       Train $\pi$ using DQN with $Reward = R(\theta)$
6:   **End for**
7:   **Return** $\pi$
---

## IV. RESULTS

### A. Deep Q Learning

The first model is trained using deep reinforcement learning alone, denoted as RL500. The model loss and reward curves are shown in Figure 2. The summary results are shown in Table 2. The model achieves an average score of 331.63 and a peak reward of 1949.39 after 500 episodes of training.

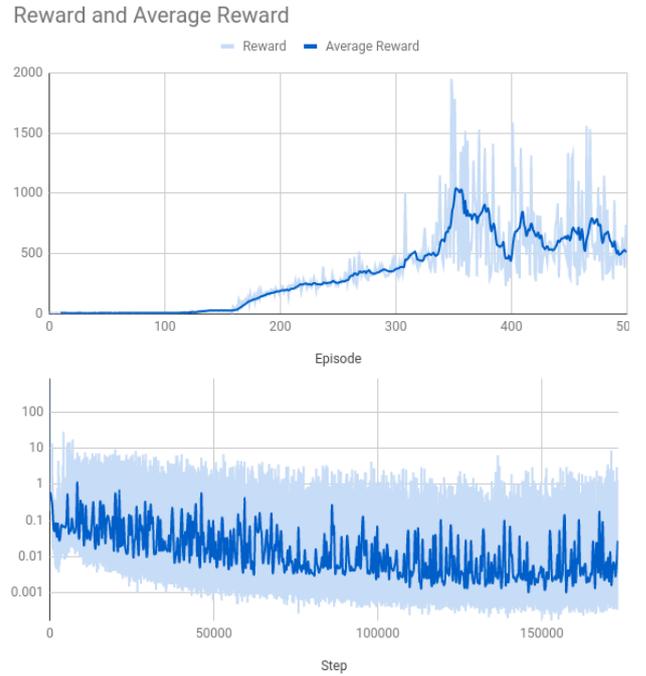

Figure 2. Model Loss and Reward for CartPole-v1 with Deep Q Learning, denoted as RL500.

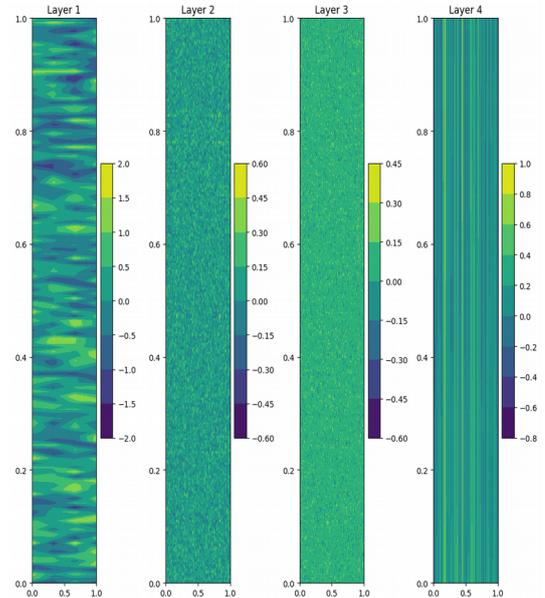

Figure 3. Model Weights for RL500 Model.

### B. Q-Imitation-Learning with Expert Player (PID)

The second and third model are trained by Q Imitation Learning via the expert PID system, referred as IL250 and IL500. Figure 4 and 6 shows the loss and reward of the expert. The model imitation loss after 250 episodes of training and 500 episodes of training are also shown in Figure 4 and 6. Table 2 shows the IL250 producing an average score of 593.1 and a peak score of 6082.91, while IL500 reaches an average score of 681.91 and peak score of 4652.33. Model weights are shown in Figure 5 and 7.

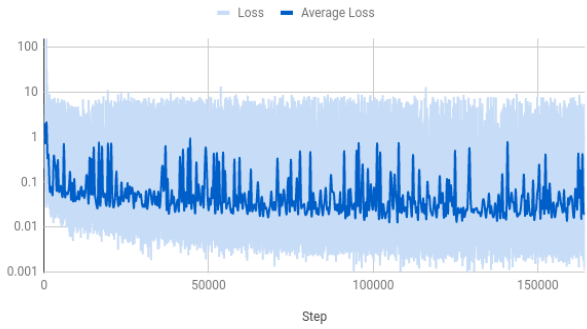

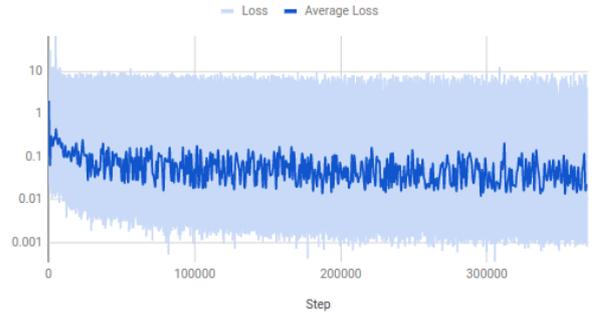

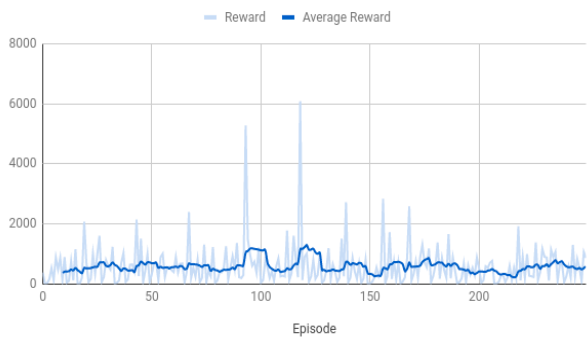

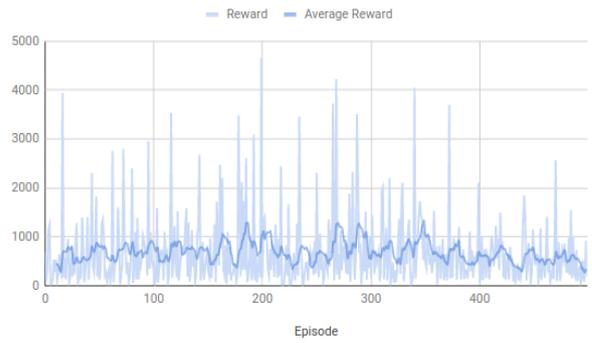

Figure 4. Model Loss and Reward for Q Imitation Learning using PID expert player, denoted as IL250.

Figure 6. Model Loss and Reward for Q Imitation Learning using PID expert player, denoted as IL500.

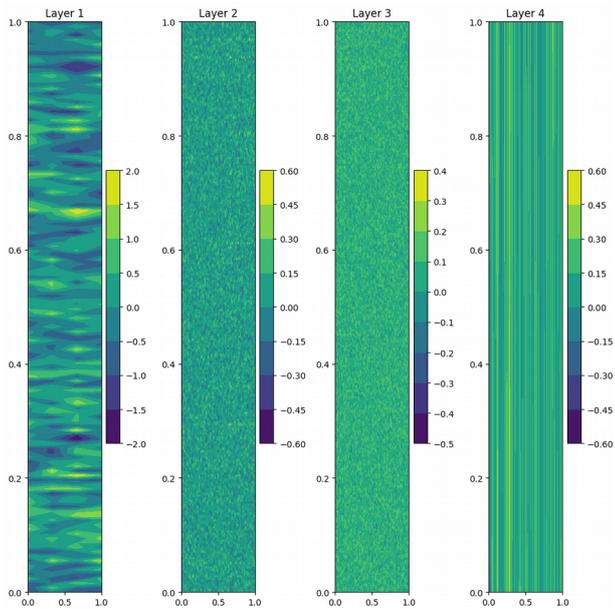

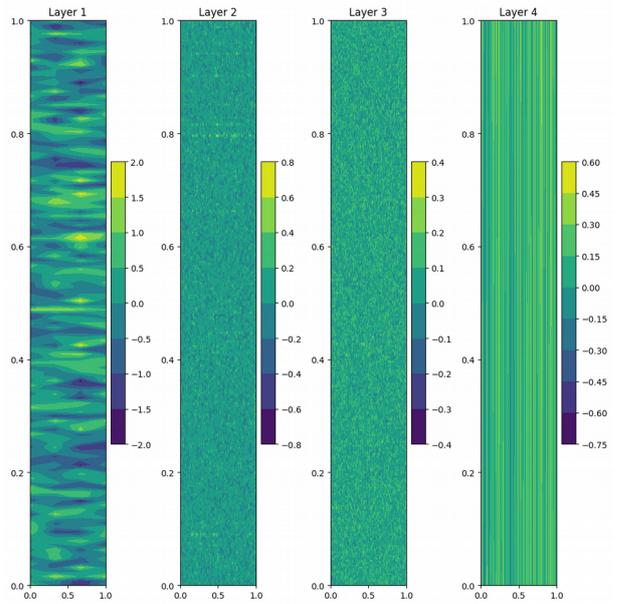

Figure 5. Model Weights for IL250 Model.

Figure 7. Model Weights for IL500 Model

## C. Augmented Q-Imitation-Learning

We train the fourth model using Deep Q Learning with the imitation learning model from B. This model is referred to as IL250 + RL250. Model loss and reward plots are shown in Figure 8. As shown in Table 2, the model achieves an average reward of 2000 and a peak reward of 13000 after 250 episodes of training. Model weights are shown in Figure 9.

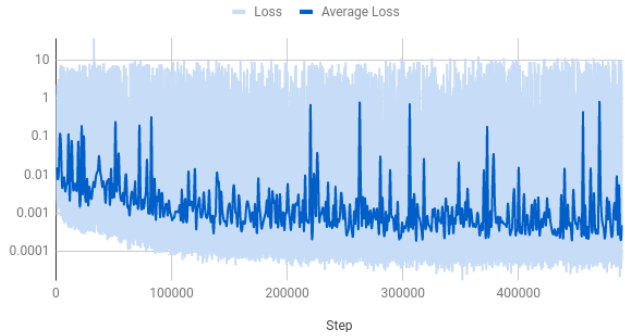

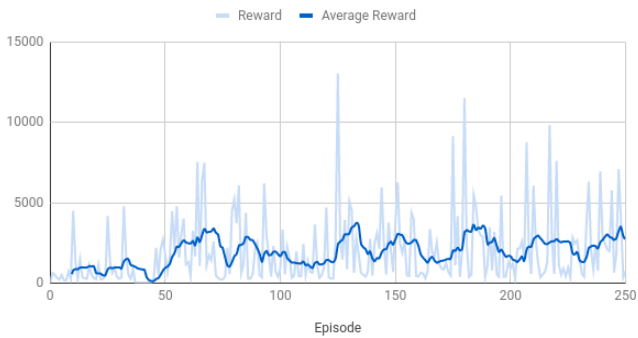

Figure 8. Model Loss and Reward for CartPole-v1 with Deep Q Learning after Q Imitation Learning, denoted as IL250 + RL250.

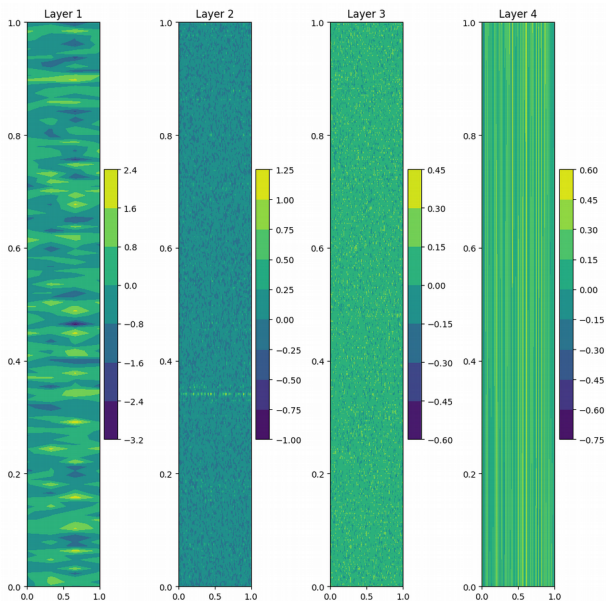

Figure 9. Model Weights for IL250 + RL250 Model.

|  | CartPole-v1 |
| --- | --- |
| RL500 | 331.63 |
| IL250 | 593.1 |
| IL500 | 681.91 |
| IL250 + RL250 | 1937.61 |
| RL500 Best | 1949.39 |
| IL250 Best | 6082.91 |
| IL500 Best | 4652.33 |
| IL250 + RL250 Best | 13043.02 |

Table 2. Comparison of average reward and best reward for each learning method. Best reward measures the single best performing episode for each learning method.

## V. CONCLUSION

IL250+RL250 outperforms IL250, IL500 and RL500 by a large margin on both average and best score. This is given that the total number of episodes are equal. This shows that AQIL may indeed be an effective approach to accelerate Q-learning training in cases where an expert system is readily available and the Deep Q-learning methodology is relevant.

Future research should include characterizing performance of AQIL with changes in training configuration, model topology and environment complexity. It would be interesting to see the performance of AQIL on a variety of environments. Repeated bouts of imitation and reinforcement learning may also provide some insight. Lastly the effects of different reward functions and error functions should be characterized.

REFERENCES

[1] Reddy, S., Dragan, A.D., & Levine, S. (2019). SQIL: Imitation Learning via Regularized Behavioral Cloning. *ArXiv, abs/1905.11108*.

[2] Attia, A., & Dayan, S. (2018). Global overview of Imitation Learning. *ArXiv, abs/1801.06503*.

[3] Judah, K., Fern, A., Dietterich, T.G., & Tadepalli, P. (2014). Active Imitation learning: formal and practical reductions to I.I.D. learning. *J. Mach. Learn. Res., 15*, 3925-3963.

[4] Ng, A.Y., & Russell, S.J. (2000). Algorithms for Inverse Reinforcement Learning. *ICML*.

[5] Haarnoja, T., Tang, H., Abbeel, P., & Levine, S. (2017). Reinforcement Learning with Deep Energy-Based Policies. *ICML*.

[6] Mnih, V., Kavukcuoglu, K., Silver, D., Graves, A., Antonoglou, I., Wierstra, D., & Riedmiller, M.A. (2013). Playing Atari with Deep Reinforcement Learning. *ArXiv, abs/1312.5602*.